\documentclass[conference]{IEEEtran}
\IEEEoverridecommandlockouts

\usepackage{cite}
\usepackage{amsmath,amssymb,amsfonts}
\usepackage{algorithmic}
\usepackage{graphicx}
\usepackage{textcomp}
\usepackage{xcolor}
\usepackage{algorithm}
\usepackage{booktabs}

\begin{document}

\title{Attention-Guided Autoencoder Fusion for Insulator
Defect Detection Using UAV Transmission-Line Imaging}

\author{\IEEEauthorblockN{Malak Allam}
\IEEEauthorblockA{\textit{Faculty of Computer Science} \\
\textit{MSA University}\\
Giza, Egypt \\
malak.abdelaziz@msa.edu.eg}

\and
\IEEEauthorblockN{Khaled Shaban}
\IEEEauthorblockA{\textit{Department of Computer Science and Engineering} \\
\textit{Qatar University}\\
Doha, Qatar \\
khaled.shaban@qu.edu.qa}

\and
\IEEEauthorblockN{Ali Hamdi}
\IEEEauthorblockA{\textit{Faculty of Computer Science} \\
\textit{MSA University}\\
Giza, Egypt \\
ahamdi@msa.edu.eg}
}

\maketitle
\thispagestyle{plain} 
\pagestyle{plain}     
\begin{abstract}
Automated defect detection in high-voltage transmission-line insulators remains challenging due to severe class imbalance, large scale variation, and the small spatial extent of defect instances in Unmanned Aerial Vehicle (UAV) imagery. To address these challenges, this paper proposes AE-YOLO, an Attention-Guided AutoEncoder-Enhanced YOLO framework for robust insulator defect detection.
The architecture integrates lightweight bottleneck autoencoders within a Feature Pyramid Network-Path Aggregation Network (FPN-PAN) neck. This preserves anomaly-sensitive information during multi-scale feature fusion. Convolutional Block Attention Modules (CBAM) are used throughout the backbone, enhancing feature discrimination and suppressing background interference. The framework also introduces a variance-maximizing autoencoder regularization strategy, which encourages diverse, defect-discriminative latent representations.
The network trains using a unified objective. This combines focal loss, Complete IoU (CIoU) loss, and autoencoder regularization to address foreground-background imbalance and improve localization accuracy. During inference, Weighted Boxes Fusion (WBF) combines predictions from YOLOv8, YOLOv10, and YOLO11. An autoencoder-guided confidence boosting mechanism simultaneously enhances sensitivity to rare defect categories.
Experiments on the Insulator-Defect Detection dataset showed AE-YOLO, with an EfficientNetV2 backbone, achieved 95.10\% mAP@0.5, 96.40\% precision, and 93.80\% recall. This performance surpassed the strongest YOLO-family baseline by 5.0 points in mAP@0.5 and 6.7 points in recall. These consistent improvements across five backbone architectures also confirmed the framework's effectiveness and adaptability. Together, these results suggest AE-YOLO could be a practical, scalable solution for UAV-based transmission-line inspection and defect monitoring.
\end{abstract}

\section{Introduction}
High-voltage transmission lines rely on insulators to provide electrical insulation and mechanical support for conductors attached to transmission towers. Typically made of glass or porcelain, these components operate under high electrical stress, mechanical loading, and harsh environmental conditions~\cite{miao2019insulator}. Over time, such conditions can lead to self-explosion, structural degradation, leakage currents, and pollution-induced flashovers~\cite{li2020insulator,wang2022self}. Among these failure modes, contamination and physical defects are the primary causes of insulation degradation and power outages. Pollution flashover occurs when contaminants accumulate on the insulator surface and form a conductive layer under humid conditions, increasing leakage currents and reducing insulation performance~\cite{zhao2021influence}. Consequently, early detection of insulator defects is essential for improving grid reliability, reducing maintenance costs, and preventing service interruptions. Recent studies have explored deep learning-based methods for early flashover prediction using image and video analysis~\cite{ottakath2026flashdetr}.

Traditional inspection methods, such as helicopter patrols, manual visual inspection, and rule-based image analysis, are labor-intensive, costly, and difficult to scale across large transmission networks~\cite{saranya2016svm,wang2019image}. The adoption of Unmanned Aerial Vehicles (UAVs) has significantly improved transmission-line inspection by enabling low-cost, high-resolution data collection from inaccessible locations while supporting near real-time analysis on edge devices~\cite{elmancy2025real}. However, UAV imagery presents challenges including small defect targets, complex backgrounds, scale variations, illumination changes, and occlusions from surrounding infrastructure or vegetation~\cite{li2021pin}.

Recent advances in deep learning have substantially improved automated defect detection. Two-stage detectors such as R-CNN, Fast R-CNN, and Faster R-CNN~\cite{girshick2014rich,girshick2015fast,ren2015faster} achieve strong localization performance and have been successfully applied to insulator inspection~\cite{lu2021insulator,liao2019power}. However, their computational complexity limits real-time deployment on UAV platforms. In contrast, one-stage detectors such as YOLO and SSD~\cite{redmon2016you,liu2016ssd} provide a better balance between accuracy and speed. Recent approaches based on YOLOv5, MTI-YOLO, and CenterNet variants have shown promising results for transmission-line inspection~\cite{feng2021electrical,liu2021mti,wu2021insulator}. Nevertheless, detecting small defects and maintaining high recall under severe class imbalance remain challenging.

To address these limitations, attention mechanisms have been increasingly integrated into object detection frameworks~\cite{allam2025attention}. Modules such as CBAM~\cite{woo2018cbam} enhance feature representations through channel and spatial attention, improving focus on informative regions while suppressing background noise. Similar strategies have demonstrated effectiveness in aerial inspection tasks, including solar panel fault detection and plant disease monitoring~\cite{elfeky2025automated,yehia2024enhanced}, highlighting their potential for detecting small and subtle defects.

Feature Pyramid Networks (FPNs) and Path Aggregation Networks (PANs) are widely used to fuse multi-scale features and improve detection across varying object sizes~\cite{lin2017feature,liu2018path}. However, these architectures primarily optimize semantic representation and may weaken anomaly-discriminative features during feature fusion, particularly when defects occupy only a small portion of the image. In addition, most detection frameworks do not leverage reconstruction-based regularization to preserve anomaly-sensitive latent representations throughout the detection pipeline.

Motivated by these challenges, this paper proposes AE-YOLO, an Attention-Guided AutoEncoder-Enhanced YOLO framework for insulator defect detection in UAV-based transmission-line inspection. AE-YOLO integrates lightweight bottleneck autoencoders within the FPN neck to preserve anomaly-discriminative information during multi-scale fusion, while CBAM modules enhance feature discrimination and suppress background clutter. A variance-maximizing autoencoder regularization strategy encourages diverse latent representations sensitive to rare defects. Finally, Weighted Boxes Fusion (WBF) and an autoencoder-guided confidence boosting mechanism improve robustness and enhance the detection of difficult and infrequent defect categories.

The main contributions of this work are summarized as follows:
\begin{itemize}
    \item We propose AE-YOLO,  a novel attention-guided and autoencoder-enhanced object detection framework that integrates lightweight bottleneck autoencoders within the FPN-PAN architecture to improve multi-scale defect representation localization.
    
    \item We introduce a backbone-agnostic design that supports multiple feature extraction architectures, including EfficientNetV2, MobileNetV3, ResNet50, DenseNet201, and ConvNeXt-Tiny, enabling deployment across a wide range of computational environments.
    
    \item We develop a unified optimization objective that combine focal loss, Complete IoU (CIoU) loss, and variance-maximizing autoencoder regularization to address class imbalance and improve sensitivity to small defect instances.

\end{itemize}

The remainder of this paper is organized as follows. Section II reviews related work on insulator defect detection and deep learning-based inspection systems. Section III describes the dataset and its characteristics. Section IV presents the proposed AE-YOLO framework. Section V reports the experimental setup and results. Section VI discusses the findings and limitations of the study, and Section VII concludes the paper and outlines future research directions.

\section{Related Work}

\begin{figure*}[!t]
\centering
\includegraphics[width=0.38\textwidth]{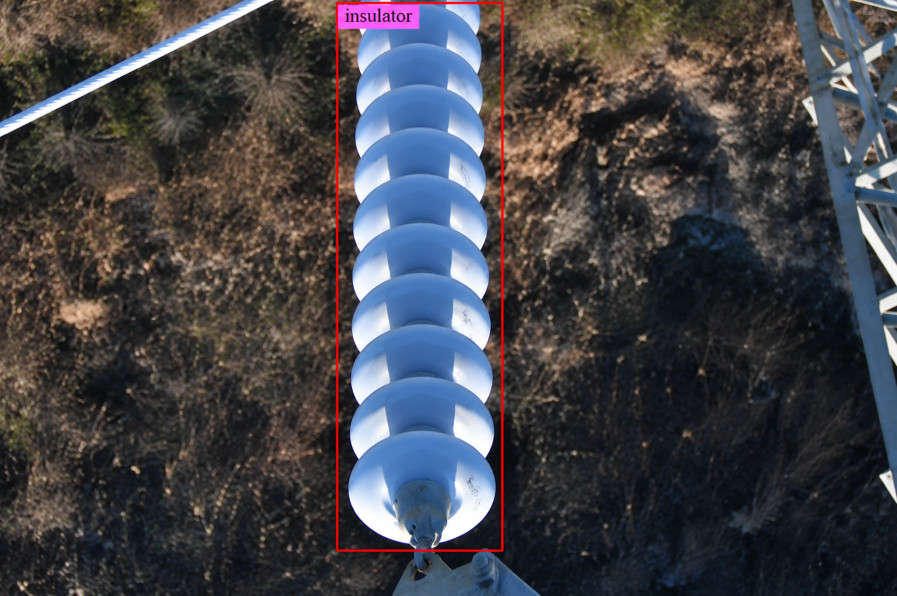}
\hfill
\includegraphics[width=0.29\textwidth]{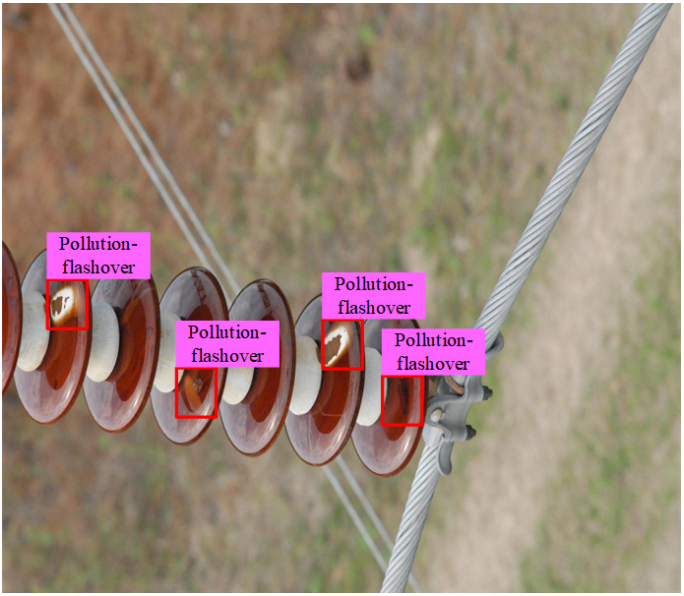}
\hfill
\includegraphics[width=0.29\textwidth]{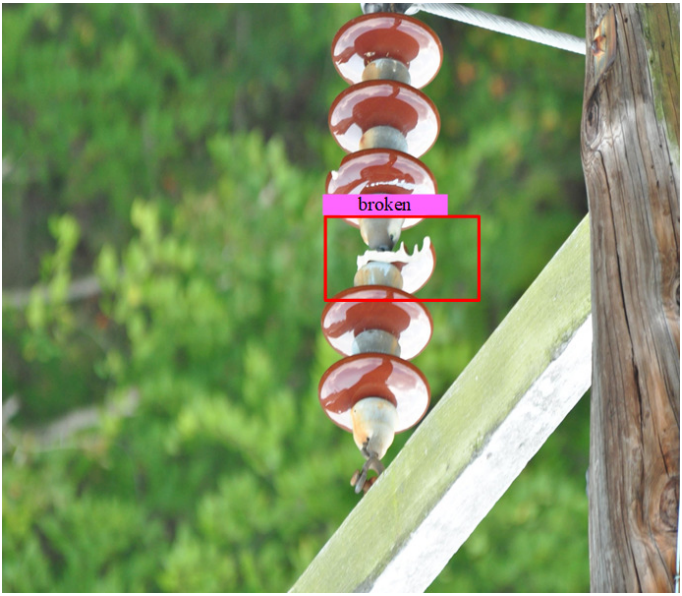}
\caption{Sample annotated images from the Insulator-Defect Detection dataset, showing ground-truth bounding boxes for insulator, pollution-flashover, and broken defect instances under varying imaging conditions.}
\label{fig:test}
\end{figure*}

\subsection{Insulator Defect Detection}
Automated insulator inspection has evolved significantly from traditional image-processing techniques to advanced deep learning-based detection frameworks. Early approaches primarily relied on hand-crafted features, thresholding methods, and rule-based image analysis. Although these methods achieved acceptable performance under controlled conditions, their robustness deteriorated considerably in the presence of complex backgrounds, illumination variations, and occlusions commonly encountered in UAV-acquired imagery~\cite{wang2019image,saranya2016svm}. Machine learning methods based on Support Vector Machines (SVMs) subsequently improved fault classification performance; however, their effectiveness remained highly dependent on manually engineered features and domain-specific preprocessing pipelines~\cite{saranya2016svm}.

The emergence of deep learning substantially improved defect detection accuracy by enabling automatic feature extraction. Two-stage object detectors, including Faster R-CNN and its variants, have demonstrated strong localization capabilities for insulator inspection tasks\cite{lu2021insulator,ren2015faster,liao2019power}. These methods typically achieve high detection accuracy through region proposal mechanisms and multi-stage refinement processes. However, their computational complexity and inference latency limit their applicability in real-time UAV inspection scenarios, where rapid decision-making is essential.

o address these limitations, researchers increasingly adopted one-stage detectors. YOLO-based frameworks and SSD-derived architectures provide a more favorable balance between accuracy and computational efficiency by directly predicting object locations and class labels in a single forward pass\cite{li2021pin,elmancy2025real,ottakath2026flashdetr,feng2021electrical}. Recent studies have further enhanced these architectures through multi-task learning, anchor-free detection strategies, and lightweight network designs\cite{feng2021electrical,liu2021mti,wu2021insulator}. Nevertheless, the reliable detection of small-scale defects remains challenging because defect regions often occupy only a tiny fraction of the image and exhibit significant visual similarity to surrounding background structures. Furthermore, severe class imbalance frequently causes conventional detectors to prioritize dominant classes at the expense of rare defect categories.

\subsection{Multi-Scale Feature Fusion for Object Detection}

Accurate detection of objects exhibiting large scale variations remains one of the fundamental challenges in computer vision. FPN introduced a top-down architecture with lateral connections that allowed the propagation of semantically rich information across multiple resolution levels\cite{lin2017feature}. This design significantly improved the detection of small objects by combining high-level semantic information with low-level spatial details.

PANs further enhanced feature fusion by introducing complementary bottom-up pathways that strengthen information flow across feature hierarchies\cite{liu2018path}. Consequently, FPN-PAN architectures have become a standard component of modern object detectors, including several generations of the YOLO family. More recently, architectures such as YOLOv7 have demonstrated that carefully designed multi-scale feature aggregation strategies can substantially improve detection accuracy while maintaining real-time performance~\cite{wang2022yolov7}.

Despite their effectiveness, conventional feature pyramid architectures are primarily optimized for semantic feature aggregation rather than anomaly preservation. During repeated feature compression and fusion operations, subtle defect-related cues may be weakened or discarded, particularly when defects occupy very small image regions. This limitation becomes especially critical in transmission-line inspection, where successful detection often depends on preserving fine-grained anomaly characteristics across multiple scales.

\subsection{Attention Mechanisms for Defect Detection}

Attention mechanisms have become an important component of modern deep learning architectures because they enable adaptive emphasis of informative features while suppressing irrelevant information. Squeeze-and-Excitation (SE) networks introduced channel-wise feature recalibration through lightweight gating operations, significantly improving representational capacity with minimal computational overhead~\cite{hu2018squeeze}. 

Building upon this concept, the CBAM integrates both channel and spatial attention mechanisms to enhance feature discrimination\cite{woo2018cbam}. By sequentially learning what information is important and where it is located, CBAM effectively improves localization performance and reduces the influence of background clutter. Such capabilities are particularly valuable in aerial inspection applications, where defect regions are often small, visually subtle, and surrounded by complex environmental structures.
 
Attention-guided architectures have demonstrated promising performance across several inspection domains, including thermal fault detection in photovoltaic systems\cite{allam2025attention}, and semantic segmentation under challenging illumination conditions~\cite{elmahdy2024rhrsegnet}. These studies consistently indicate that attention mechanisms enhance feature quality and improve robustness to environmental variability. However, most existing approaches employ attention solely for feature enhancement and do not explicitly incorporate mechanisms that preserve anomaly-specific latent representations during multi-scale feature fusion.

\subsection{Efficient Backbone Architectures}

The design of efficient backbone networks has received significant attention due to the growing demand for deploying deep learning models on resource-constrained platforms. MobileNets architectures introduced depthwise separable convolutions to substantially reduce computational complexity while maintaining competitive accuracy~\cite{howard2017mobilenets,sandler2018mobilenetv2}. EfficientNet further improved efficiency through compound scaling of network depth, width, and input resolution, establishing a favorable accuracy-to-computation trade-off across multiple benchmark tasks~\cite{tan2019efficientnet}. 



More recently, DenseNet, ConvNeXt, and EfficientNetV2 architectures have demonstrated strong representational capabilities while maintaining practical computational requirements. These networks provide diverse design philosophies, ranging from dense feature reuse to modernized convolutional architectures inspired by Transformer design principles. The availability of these complementary backbone architectures motivates the development of backbone-agnostic detection frameworks capable of utilizing their respective strengths without requiring substantial architectural modifications.

\subsection{Autoencoders for Anomaly-Aware Feature Learning}

Autoencoders have been widely used for unsupervised anomaly because of their ability to learn compact representations of normal data distributions~\cite{hinton2006reducing}. By minimizing reconstruction error during training, autoencoders develop latent representations that capture dominant structural patterns while remaining sensitive to anomalous inputs. Consequently, abnormal samples often produce elevated reconstruction errors or distinctive latent-space activations.

Beyond anomaly detection, recent studies have shown that reconstruction-based objectives can act as effective regularization mechanisms for deep neural networks. Feature-level regularization encourages latent representations to remain informative and diverse, thereby reducing the risk of feature collapse and improving generalization performance~\cite{zhao2021influence}. This property is particularly relevant for defect detection tasks involving rare classes, where preserving anomaly-discriminative information is essential for maintaining high recall.

However, most existing object detection frameworks employ autoencoders as independent anomaly detection modules rather than integrating them directly into multi-scale feature extraction pipelines. To the best of our knowledge, limited research has investigated the use of lightweight bottleneck autoencoders as feature-level regularizers within FPN-PAN architectures for UAV-based insulator defect detection.

\subsection{Loss Functions for Imbalanced Defect Detection}

Class imbalance represents a persistent challenge in defect detection datasets, where defective instances are often significantly outnumbered by background regions and normal objects. Focal loss addresses this problem by reducing the influence of easily classified samples and concentrating learning on hard examples~\cite{lin2017focal}.

For localization, CIoU loss extends traditional IoU-based objectives by jointly considering overlap area, center-point distance, and aspect-ratio consistency~\cite{zheng2020distance}. Compared with conventional regression losses, CIoU provides richer optimization signals and generally produces more accurate bounding-box predictions.

Although focal loss and CIoU loss effectively address classification imbalance and localization accuracy, respectively, they do not explicitly encourage the preservation of anomaly-sensitive latent representations. This observation motivates the integration of an additional autoencoder-based regularization objective capable of maintaining discriminative feature diversity throughout the detection process.

\section{Dataset}

\subsection{Overview}

The proposed AE-YOLO framework was evaluated using the Insulator-Defect Detection dataset~\cite{zheng2022insulator}, a publicly available benchmark specifically developed for object detection in high-voltage transmission-line inspection scenarios. Released through a collaboration between Changzhou University and the Chinese Academy of Sciences, the dataset contains 1,600 UAV-acquired images and 5,373 annotated object instances distributed across three classes: insulator, pollution-flashover, and broken. Figure~\ref{fig:test} presents representative examples from the dataset, illustrating the diversity of defect types, object scales, viewing angles, and environmental conditions. 

Each image contains at least one annotated object, resulting in a fully labeled dataset with no unlabeled samples. To ensure a consistent and unbiased evaluation protocol, the dataset was divided into three mutually exclusive subsets comprising 1,296 training images, 144 validation images, and 160 test images, corresponding to an approximate 81:9:10 training-validation-test split. This partitioning enables reliable model development, hyperparameter tuning, and performance assessment.

\begin{figure*}[t]
    \centering
    \includegraphics[width=\textwidth]{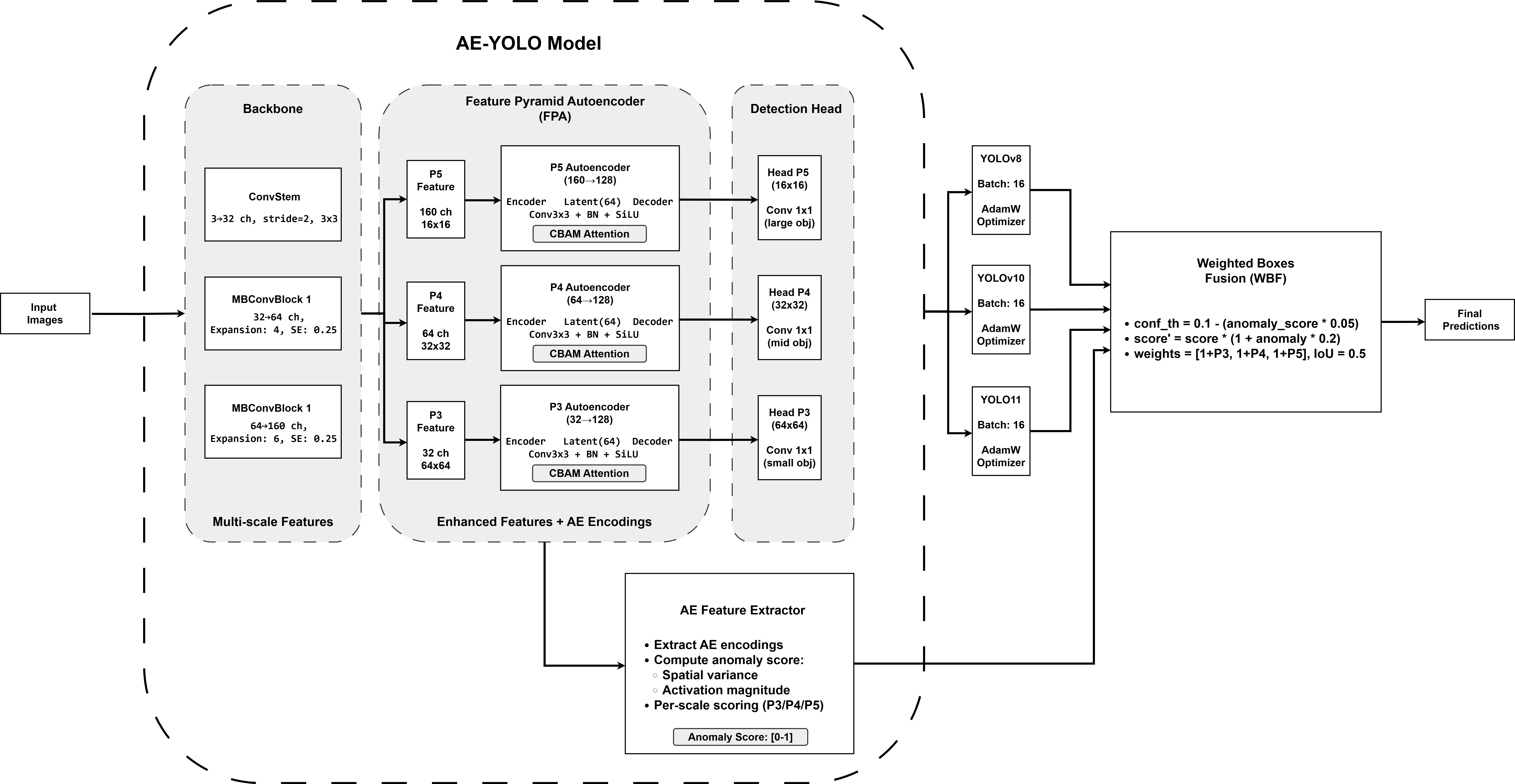}
    \caption{Overall architecture of the proposed AE-YOLO framework. The illustrated configuration employs an EfficientNetV2 backbone to generate three feature scales (P3, P4, P5), followed by a Feature Pyramid Autoencoder (FPA) neck that enhances each scale through lightweight encoder–decoder modules with CBAM attention, a multi-scale detection head, and an ensemble stage combining YOLOv8, YOLOv10, and YOLO11 via Weighted Boxes Fusion (WBF) with AE-guided confidence scoring.}
    \label{fig:architecture}
\end{figure*}
\subsection{Class Distribution and Dataset Imbalance}

Table~\ref{tab:class_stats} summarizes the class distribution and object statistics of the dataset. The analysis reveals substantial imbalance in both object frequency and spatial extent, highlighting the challenges associated with automated insulator defect detection.

The insulator class represents the dominant object category and appears in all 1,600 images, accounting for 1,832 annotated instances. Insulators exhibit significant scale variation, ranging from small distant objects to large close-up structures occupying a substantial portion of the image. Such variability necessitates the use of effective multi-scale feature extraction and fusion mechanisms capable of maintaining robust representations across different spatial resolutions.

The pollution-flashover class constitutes the most frequent defect category, with 2,448 instances distributed across 806 images. Despite its relatively high occurrence frequency, pollution-flashover defects occupy only a very small fraction of the image area, with some instances measuring as little as 26$\times$29~pixels. The extremely limited spatial footprint of these defects makes accurate localization particularly challenging and increases the likelihood of missed detections.

The broken class contains 1,093 instances across 785 images and exhibits characteristics similar to pollution-flashover defects. Broken insulator regions typically appear as small, localized anomalies that can be easily confused with background textures, shadows, or contamination artifacts. This visual similarity further complicates the discrimination between defective and non-defective regions, especially in aerial inspection imagery containing complex environmental backgrounds.

\begin{table}[t]
\centering
\caption{Per-Class Statistics of the Insulator-Defect Detection Dataset}
\label{tab:class_stats}
\resizebox{\columnwidth}{!}{%
\begin{tabular}{lccccc}
\hline
\textbf{Class} & \textbf{Images} & \textbf{Objects} &
\textbf{Obj./Img.} & \textbf{Avg.\ Area} & \textbf{Min Area} \\
\hline
Insulator           & 1600 & 1832 & 1.15 & 18.58\% & 0.46\% \\
Pollution-flashover & 806  & 2448 & 3.04 & 0.22\%  & 0.01\% \\
Broken              & 785  & 1093 & 1.39 & 0.59\%  & 0.03\% \\
\hline
\textbf{Total}      & 1600 & 5373 & --   & --      & --     \\
\hline
\end{tabular}%
}
\end{table}

\subsection{Dataset Challenges and Design Implications}

The Insulator-Defect Detection dataset presents several challenges that directly motivate the design of the proposed AE-YOLO framework.

\begin{enumerate}
    \item \textbf{Severe Class Imbalance:} Although pollution-flashover defects appear more frequently than insulator instances, their average spatial coverage is dramatically smaller. The large disparity between dominant object regions and small defect areas creates a substantial foreground-background imbalance that can bias conventional training objectives toward easily classified samples. To address this issue, AE-YOLO employs focal loss~\cite{lin2017focal}, which adaptively down-weights well-classified examples and concentrates learning on difficult defect instances.
    \item \textbf{Extreme Scale Variation:} The dataset exhibits considerable variation in object size. Insulator instances range from small aerial targets to structures occupying a large fraction of the image, while defect regions frequently appear at only a few dozen pixels in width. Such scale diversity requires the preservation of both high-resolution spatial details and high-level semantic information throughout the detection pipeline. Consequently, AE-YOLO incorporates an FPN-PAN architecture augmented with FPA modules to enhance multi-scale feature representations.
    \item \textbf{Small Defect Localization:} The accurate localization of tiny defects remains one of the most challenging aspects of transmission-line inspection. Small errors in bounding-box prediction can lead to substantial reductions in localization accuracy when defect regions occupy only a limited number of pixels. To improve localization robustness, AE-YOLO employs CIoU loss~\cite{zheng2020distance}, which jointly considers overlap area, center-point distance, and aspect-ratio consistency during bounding-box optimization.
    \item \textbf{Complex Environmental Conditions}: In addition to object-scale and class-distribution challenges, UAV-acquired inspection imagery often contains varying illumination conditions, background clutter, occlusions, and changes in viewing perspective. These factors introduce significant intra-class variability and increase the difficulty of distinguishing subtle defect patterns from surrounding structures. The integration of attention-guided feature extraction and anomaly-aware feature regularization within AE-YOLO is specifically intended to improve robustness under such challenging operating conditions.
\end{enumerate}

Overall, the characteristics of the Insulator-Defect Detection dataset provide a realistic and demanding benchmark for evaluating automated inspection systems. The combination of severe class imbalance, extreme scale variation, small defect instances, and complex environmental backgrounds makes the dataset particularly suitable for assessing the effectiveness of the proposed AE-YOLO framework.

\section{Methodology}

This section presents AE-YOLO, an Attention-Guided Autoencoder-Enhanced YOLO framework designed for robust insulator defect detection in UAV-based transmission-line inspection imagery. The proposed architecture addresses three major challenges inherent to the target application: severe class imbalance, extreme object-scale variation, and the accurate localization of small defect instances.

As illustrated in Fig.~\ref{fig:architecture}, AE-YOLO consists of four principal components: (i) a configurable attention-enhanced backbone for multi-scale feature extraction, (ii) an FPA neck for anomaly-aware feature refinement, (iii) a multi-scale detection head for object localization and classification, and (iv) an ensemble inference module employing AE-guided confidence boosting and WBF. The overall architecture is designed to preserve anomaly-discriminative information throughout the feature hierarchy while maintaining compatibility with multiple backbone configurations.

\subsection{Configurable Attention-Enhanced Backbone}

The backbone network is responsible for extracting hierarchical visual representations from input UAV images. To ensure flexibility across different deployment scenarios, AE-YOLO supports multiple backbone architectures, including EfficientNetV2, MobileNetV3, ResNet50, DenseNet201, and ConvNeXt-Tiny. Regardless of the selected backbone, three feature maps corresponding to progressively lower spatial resolutions are extracted and denoted as P3, P4, and P5.

To improve feature discrimination, CBAMs~\cite{woo2018cbam} are incorporated after each downsampling stage. CBAM sequentially applies channel and spatial attention operations, enabling the network to emphasize defect-relevant regions while suppressing irrelevant background information. This capability is particularly important in UAV inspection imagery, where defect regions frequently occupy only a small fraction of the image and are surrounded by visually complex structures.

Figure\ref{fig:architecture} illustrates the the EfficientNetV2 implementation of the proposed framework. The network begins with a convolutional stem followed by a sequence of MBConv blocks equipped with Squeeze-and-Excitation (SE) modules\cite{hu2018squeeze} and SiLU activation functions. Intermediate feature maps generated at different depths are used to construct the feature pyramid and serve as inputs to the proposed Feature Pyramid Autoencoder neck.

For the EfficientNetV2 configuration, the stem consists of a 3 $\times$ 3 convolution with stride 2 followed by batch normalization and SiLU activation. Subsequently, four stages of MBConv blocks progressively increase the receptive field while reducing spatial resolution. The outputs of Stages 2, 3, and 4 correspond to the P3, P4, and P5 feature levels used throughout the remainder of the detection pipeline.

The extracted multi-scale features are subsequently refined by the proposed Feature Pyramid Autoencoder modules, which constitute the central contribution of this work.


\subsection{Feature Pyramid Autoencoder (FPA) Neck}

The main innovation of this work lies in its neck design. Unlike standard FPNs~\cite{lin2017feature}, which simply merge multi-scale features through interpolation and concatenation, our neck embeds a dedicated bottleneck autoencoder at each pyramid level. This makes the feature representations sensitive to anomalies. An encoder, trained to compress and reconstruct normal feature distributions, will generate high-activation, high-variance encodings when it encounters defective regions that don't match the expected distribution~\cite{hinton2006reducing}.

\subsubsection{Lateral Projections}

Three $1\times1$ lateral convolutions project the backbone outputs
P3 (32~ch), P4 (64~ch), and P5 (160~ch) to a uniform dimensionality
of 128 channels, decoupling backbone width from neck width and
enabling shared autoencoder design.

\subsubsection{Feature Pyramid Autoencoder (FPA) Modules}

Each pyramid level is processed by a \textit{FeaturePyramidAutoencoder}
(FPA) module. Given an input feature map $\mathbf{F} \in
\mathbb{R}^{B \times C \times H \times W}$ with $C = 128$ channels,
the module operates as follows.

The \textit{encoder} applies a $1\times1$ convolution to compress
$\mathbf{F}$ to a bottleneck dimension of $C_b = \lfloor C \cdot r
\rfloor$ channels, where the bottleneck ratio $r = 0.25$ yields
$C_b = 32$, followed by batch normalization and SiLU activation:
\begin{equation}
    \mathbf{Z} = \text{SiLU}\bigl(\text{BN}\bigl(\text{Conv}_{1\times1}
    (\mathbf{F})\bigr)\bigr), \quad \mathbf{Z} \in
    \mathbb{R}^{B \times C_b \times H \times W}.
\end{equation}

A channel attention gate is then applied to the bottleneck
representation. Global average pooling collapses spatial dimensions,
and a two-layer $1\times1$ MLP with ReLU and Sigmoid activations
(bottleneck factor~4) produces per-channel weights:
\begin{equation}
    \mathbf{A} = \sigma\!\left(\text{Conv}_{1\times1}\!\left(
    \text{ReLU}\!\left(\text{Conv}_{1\times1}
    \!\left(\text{GAP}(\mathbf{Z})\right)\right)\right)\right),
\end{equation}
yielding attended encoding $\hat{\mathbf{Z}} = \mathbf{Z} \odot
\mathbf{A}$.

The \textit{decoder} reconstructs the full channel dimension via
a further $1\times1$ convolution and batch normalization:
\begin{equation}
    \hat{\mathbf{F}} = \text{BN}\bigl(\text{Conv}_{1\times1}
    (\hat{\mathbf{Z}})\bigr), \quad \hat{\mathbf{F}} \in
    \mathbb{R}^{B \times C \times H \times W}.
\end{equation}

The enhanced feature is formed by a learnable residual:
\begin{equation}
    \mathbf{F}^{+} = \mathbf{F} + \alpha \cdot \hat{\mathbf{F}},
\end{equation}
where $\alpha$ is a scalar parameter initialized to 0.1 and learned
end-to-end. This initialization ensures training stability by starting
from an identity mapping and gradually incorporating the autoencoder
signal as it becomes reliable.

\subsubsection{Top-Down and Bottom-Up Pathways}

After per-level AE enhancement, features combine in an FPN-PAN~\cite{liu2018path} topology. In the top-down pass, P5 is upsampled to P4's spatial resolution through nearest-neighbor interpolation. This is then element-wise added to the lateral P4 projection and smoothed with a $3\times3$ convolution. The result then goes through the P4 FPA module. This process repeats for P4 to P3. For the bottom-up PAN pass, P3 gets downsampled via a strided $3\times3$ convolution. Then, it's concatenated with the enhanced P4 and projected to 128 channels using a $1\times1$ fusion layer. Similar steps aggregate P4 into P5. This two-way pathway ensures both fine spatial details and strong semantic context are available across all three detection scales.

\subsection{Multi-Scale Detection Head}

A single shared detection head is applied independently to each of
the three enhanced pyramid levels. For each scale, the head applies
a $3\times3$ depthwise-separable-style convolution followed by
batch normalization and SiLU, then a $1\times1$ projection to
$3 \times (N_c + 5)$ output channels, where $N_c$ is the number of
target classes and the factor 3 corresponds to the number of anchors
per scale. The three output tensors correspond to large objects at
P5 ($16 \times 16$ grid at $512^2$ input), medium objects at P4
($32 \times 32$), and small objects at P3 ($64 \times 64$),
providing detection coverage across the full scale range present
in the insulator dataset.

\subsection{Training Objective}

The total training loss $\mathcal{L}$ combines three components:
\begin{equation}
    \mathcal{L} = \lambda_{\text{cls}}\,\mathcal{L}_{\text{cls}}
    + \lambda_{\text{box}}\,\mathcal{L}_{\text{box}}
    + \lambda_{\text{ae}}\,\mathcal{L}_{\text{ae}},
\end{equation}
with weights $\lambda_{\text{cls}} = 0.5$,
$\lambda_{\text{box}} = 7.5$, and $\lambda_{\text{ae}} = 1.0$.

\textbf{Classification loss:} Focal loss~\cite{lin2017focal} with
$\alpha = 0.25$ and $\gamma = 2.0$ is used for the binary
classification head at each anchor:
\begin{equation}
    \mathcal{L}_{\text{cls}} = -\alpha_t
    (1 - p_t)^{\gamma} \log p_t,
\end{equation}
where $p_t$ is the predicted probability for the ground-truth class.
This formulation down-weights easy negatives, which dominate the
training signal given the significant background-to-defect ratio
in the dataset.

\textbf{Localization loss:} Complete IoU (CIoU)
loss~\cite{zheng2020distance} is used for bounding-box regression:
\begin{equation}
    \mathcal{L}_{\text{box}} = 1 - \text{IoU}
    + \frac{\rho^2(\mathbf{b}, \mathbf{b}^{gt})}{c^2}
    + \alpha_v \cdot v,
\end{equation}
where $\rho^2(\mathbf{b}, \mathbf{b}^{gt})$ is the squared
Euclidean distance between predicted and ground-truth box centers,
$c$ is the diagonal of the smallest enclosing box, $v$ measures
aspect-ratio consistency, and $\alpha_v$ is a trade-off weight.

\textbf{Autoencoder regularization loss:} To prevent the bottleneck
encodings from collapsing to trivial constant representations, a
variance-maximizing regularization term is applied over all three
FPA encodings $\{\mathbf{Z}_k\}_{k=1}^{3}$:
\begin{equation}
    \mathcal{L}_{\text{ae}} = \frac{1}{K} \sum_{k=1}^{K}
    \Bigl(-\lambda_v \bigl(\text{Var}_s(\mathbf{Z}_k)
    + \text{Var}_b(\mathbf{Z}_k)\bigr)
    + \lambda_r \|\mathbf{Z}_k\|_2^2\Bigr),
\end{equation}
where $\text{Var}_s$ and $\text{Var}_b$ denote variance across
spatial positions and batch elements respectively,
$\lambda_v = 0.01$, and $\lambda_r = 0.001$. This encourages
diverse, defect-discriminative latent codes while preventing
activation explosion through $L_2$ regularization.

\subsection{Ensemble Inference with AE-Guided Confidence Boosting}

At inference time, three independently trained detectors, YOLOv8, YOLOv10, and YOLO11, combine with the trained AE-YOLO feature extractor in a two-stage ensemble pipeline.

\subsubsection{Anomaly Score Computation}

Given a test image, the AE-YOLO feature extractor computes a
per-image anomaly score $s$ from the three FPA bottleneck
encodings. For each encoding $\mathbf{Z}_k$, the product of
spatial variance and mean absolute activation is computed:
\begin{equation}
    s_k = \text{Var}_{hw}(\mathbf{Z}_k) \cdot
    \mathbb{E}_{hw}\!\left[|\mathbf{Z}_k|\right],
\end{equation}
and the total anomaly score is $s = \frac{1}{K}\sum_k s_k$.
Images containing prominent defects produce high spatial variance
in the bottleneck because the compressed representation must
encode the anomalous pattern; clean images produce low, spatially
uniform encodings.

\subsubsection{Confidence Boosting}

The anomaly score is converted into a multiplicative confidence
boost factor $\beta$ applied to all detection scores prior to
fusion:
\begin{equation}
    \beta = 1 + \min\!\left(0.25,\; 0.15 \cdot s\right),
    \quad \beta \in [1.0, 1.25].
\end{equation}
This mechanism increases detection recall on high-anomaly images
while leaving confident, clean-background predictions largely
unaffected.

\subsubsection{Weighted Boxes Fusion}
We use Weighted Boxes Fusion (WBF)~\cite{solovyev2021weighted} to combine predictions from all three detectors, covering both standard inference and test-time augmentation (TTA) results. Unlike Non-Maximum Suppression (NMS), which simply suppresses overlapping boxes, WBF calculates their confidence-weighted averages. We assign each model's predictions a weight proportional to its validation mAP50, then normalize these weights to sum to one. We set the fusion IoU threshold at 0.5 and discard any boxes with a post-fusion confidence below 0.25. This strategy often outperforms single-model NMS, especially for densely packed small defects where aggressive suppression often leads to missed detections.

\section{Experimental Results}

\subsection{Experimental Setup}

All experiments were conducted using the Insulator-Defect Detection dataset described in Section III. The dataset was partitioned into training, validation, and test subsets comprising 1,296, 144, and 160 images, respectively. Input images were resized to 640 $\times$ 640 pixels and processed using standard data augmentation techniques, including random horizontal flipping, scaling, and color-space perturbations.

To evaluate the effectiveness of the proposed framework, AE-YOLO was implemented using five backbone architectures: EfficientNetV2, MobileNetV3, ResNet50, DenseNet201, and ConvNeXt-Tiny. Performance was assessed using standard object detection metrics, including Precision, Recall, mAP@0.5, and mAP@0.5:0.95.

For fair comparison, all baseline YOLO-family models were trained using identical dataset partitions, image resolutions, and optimization settings. This ensures that observed performance differences are attributable to architectural modifications rather than experimental configuration variations.

\subsection{Quantitative Performance Comparison}

Table~\ref{tab:prior} presents the performance comparison between AE-YOLO and the corresponding baseline YOLO models across different backbone configurations.

The results demonstrate that AE-YOLO consistently outperforms all baseline models across every evaluation metric and backbone architecture. This consistency indicates that the proposed architectural enhancements provide generalizable performance gains rather than improvements limited to a specific feature extractor.

Among all evaluated configurations, AE-YOLO equipped with the EfficientNetV2 backbone achieved the strongest overall performance, reaching 96.40\% precision, 93.80\% recall, and 95.10\% mAP@0.5. Compared with the strongest baseline detector, this corresponds to an improvement of approximately 5.0 percentage points in mAP@0.5 and 6.7 percentage points in recall.

The largest improvements are observed in recall, suggesting that the proposed framework is particularly effective at recovering difficult defect instances that conventional detectors fail to identify. This behavior is consistent with the intended role of the Feature Pyramid Autoencoder modules, which preserve anomaly-discriminative information during multi-scale feature fusion and improve sensitivity to subtle defect characteristics.

Furthermore, performance improvements are observed across lightweight and high-capacity backbone architectures alike. The consistent gains achieved by MobileNetV3, EfficientNetV2, ResNet50, DenseNet201, and ConvNeXt-Tiny demonstrate that the effectiveness of AE-YOLO is largely independent of the underlying feature extractor, highlighting the modularity and scalability of the proposed design.

\begin{table}[t]
\centering
\caption{Comparison with Prior Insulator Defect Detection Methods}
\label{tab:prior}
\resizebox{\columnwidth}{!}{%
\begin{tabular}{lcc}
\hline
\textbf{Method} & \textbf{mAP@0.5 (\%)} & \textbf{Recall (\%)} \\
\hline
Faster R-CNN~\cite{ren2015faster}   & 86.2 & 82.1 \\
YOLOv5s                             & 89.8 & 84.5 \\
YOLOv7       & 90.1 & 87.1 \\
Improved YOLOv7~\cite{zheng2022insulator} & 93.8 & 93.4 \\
\hline
\textbf{AE-YOLO + EfficientNetV2 (Ours)} & \textbf{95.10} & \textbf{93.80} \\
\hline
\end{tabular}}
\end{table}

\subsection{Comparison with Contemporary Baseline Models}

Table~\ref{tab:results} shows how well five current YOLO-family baselines and all proposed AE-YOLO variants detect objects.

YOLOv7 stands out among the baselines. It hit an mAP@0.5 of 90.10\%, a Precision of 93.60\%, and a Recall of 87.10\%. This really points to the strength of its re-parameterized modules and trainable bag-of-freebies strategy~\cite{wang2022yolov7}. YOLOv8 and YOLO11 weren't far off, scoring mAP@0.5 values of 89.30\% and 89.40\%. But they did show much lower recall (83.20\% and 83.00\%). That suggests these models aren't as good at picking up smaller or rarer defects, like pollution-flashover. Bringing up the rear were YOLOv10 and YOLO26, with mAP@0.5 scores of 86.10\% and 87.20\%. Their poor recall means they struggle to consistently find minority defect categories when faced with complex aerial backgrounds.

\subsection{Performance of Proposed AE-YOLO Variants}

Even the lightest AE-YOLO setup, AE-YOLO + MobileNetV3, manages a strong mAP50 of 89.60\% with 87.20\% recall. It matches or even surpasses the recall of all baseline models, despite its smaller computational footprint. This performance suggests the autoencoder-regularized FPN neck effectively enhances features, regardless of the backbone's capacity.

Stronger backbones consistently improve all metrics. For instance, AE-YOLO + ResNet50 hits 90.80\% mAP50, outperforming the top baseline (YOLOv7 at 90.10\%) and boosting recall to 88.40\%. AE-YOLO + DenseNet201 elevates performance further, reaching 92.70\% mAP@0.5, 94.50\% Precision, and 90.10\% Recall. This indicates DenseNet’s dense feature reuse pairs particularly well with bottleneck autoencoder regularization for finding defects across various scales.

Our two highest-capacity configurations delivered the strongest results.
AE-YOLO + ConvNeXt-Tiny achieved 94.00\% mAP50, with 95.20\% Precision and 92.30\% Recall. AE-YOLO + EfficientNetV2, however, stood out, performing best overall across all metrics: 95.10\% mAP50, 96.40\% Precision, and 93.80\% Recall. Compared to the best baseline (YOLOv7), we saw a 5.0 percentage-point gain in mAP@0.5. Recall also improved by 6.7 percentage points. This specific metric proved especially sensitive to the rare, small-scale defect instances that make the dataset so challenging.

\begin{table}[t]
\centering
\caption{Detection Performance of Baseline and Proposed Models on the
         Insulator-Defect Detection Test Set}
\label{tab:results}
\resizebox{\columnwidth}{!}{%
\begin{tabular}{lccc}
\hline
\textbf{Model} & \textbf{mAP50 (\%)} & \textbf{Precision (\%)} & \textbf{Recall (\%)} \\
\hline
\multicolumn{4}{l}{\textit{Baseline Models}} \\
\hline
YOLOv7              & 90.10 & 93.60 & 87.10 \\
YOLOv8              & 89.30 & 92.90 & 83.20 \\
YOLOv10             & 86.10 & 84.40 & 83.70 \\
YOLO11              & 89.40 & 92.00 & 83.00 \\
YOLO26              & 87.20 & 87.00 & 81.90 \\
\hline
\multicolumn{4}{l}{\textit{Proposed AE-YOLO Variants}} \\
\hline
AE-YOLO + MobileNetV3    & 89.60 & 91.80 & 87.20 \\
AE-YOLO + ResNet50       & 90.80 & 93.10 & 88.40 \\
AE-YOLO + DenseNet201    & 92.70 & 94.50 & 90.10 \\
AE-YOLO + ConvNeXt-Tiny  & 94.00 & 95.20 & 92.30 \\
AE-YOLO + EfficientNetV2 & \textbf{95.10} & \textbf{96.40} & \textbf{93.80} \\
\hline
\end{tabular}}
\end{table}

\subsection{Analysis}

All AE-YOLO variants consistently improve recall compared to baseline models. This suggests autoencoder regularization in the FPN neck keeps the anomaly-discriminative feature structure intact. Standard detection necks, though, often suppress these features. The variance-maximizing bottleneck loss is especially useful for the \textit{pollution-flashover} and \textit{broken} classes. Their small target size and rare instances make it hard for standard detectors to stay highly sensitive.

AE-YOLO can boost both Precision and Recall simultaneously. It does this by combining focal loss, CIoU loss, and autoencoder regularization into a single training objective. Baseline models just don't handle this trade-off as well.

\section{Discussion}

The experimental results collectively demonstrate that AE-YOLO effectively addresses the principal challenges associated with UAV-based insulator defect detection. The integration of attention-guided feature extraction, autoencoder-regularized multi-scale fusion, and anomaly-aware ensemble inference enables the framework to achieve superior detection performance across diverse backbone architectures.

Particularly noteworthy is the substantial improvement in recall, which is often the most critical metric in infrastructure inspection applications. Missing a defective insulator may lead to costly maintenance failures or safety risks, whereas a limited increase in false positives can typically be addressed through secondary inspection procedures. Therefore, the observed recall improvements highlight the practical value of the proposed framework for real-world transmission-line monitoring systems.

For deployments in resource-constrained environments, like onboard drones, the MobileNetV3 and ResNet50 variants already outperform all baseline recall numbers, and at a lower computational cost. They offer a practical option that's less computationally expensive than the full EfficientNetV2 setup.

A few limitations remain, though. We didn't benchmark AE-YOLO's inference speed, so directly comparing its latency to previous methods is difficult. Also, since we only evaluated it on one dataset, we can't yet confirm if it generalizes to other insulator types and environments. Finally, we haven't independently tested the contributions of CBAM attention~\cite{woo2018cbam} and transformer-based cross-scale fusion. Structured ablation experiments will be needed to figure out what each architectural component does on its own.

Overall, the results demonstrate that AE-YOLO provides a robust, scalable, and accurate solution for automated insulator defect detection and establishes a strong foundation for future research on anomaly-aware object detection in aerial inspection applications.

\section{Conclusion}

This paper presented AE-YOLO, an Attention-Guided Autoencoder-Enhanced YOLO framework for automated insulator defect detection in UAV-based transmission-line inspection imagery. The proposed framework was developed to address three fundamental challenges encountered in aerial infrastructure inspection: severe class imbalance, extreme object-scale variation, and the accurate localization of small defect instances.

To overcome these challenges, AE-YOLO integrates attention-enhanced feature extraction with a novel FPA neck that preserves anomaly-discriminative information during multi-scale feature fusion. The framework further incorporates variance-maximizing latent regularization, anomaly-aware confidence boosting, and ensemble prediction fusion to improve sensitivity to rare and visually subtle defect categories.

Experimental evaluation on the Insulator-Defect Detection dataset demonstrated that the proposed approach consistently outperforms contemporary YOLO-family baselines across all evaluated backbone configurations. The best-performing model, AE-YOLO with an EfficientNetV2 backbone, achieved 95.10\% mAP@0.5, 96.40\% precision, and 93.80\% recall, surpassing the strongest baseline by 5.0 percentage points in mAP@0.5 and 6.7 percentage points in recall. These improvements were particularly pronounced for the pollution-flashover and broken defect classes, highlighting the effectiveness of the proposed anomaly-aware feature learning strategy for detecting small and infrequent defects.

The consistent performance gains observed across EfficientNetV2, MobileNetV3, ResNet50, DenseNet201, and ConvNeXt-Tiny backbones demonstrate the flexibility and backbone-agnostic nature of the proposed architecture. This characteristic enables AE-YOLO to be adapted to a wide range of deployment scenarios, from resource-constrained edge devices to high-performance inspection platforms.

From a practical perspective, the proposed framework offers a scalable and reliable solution for intelligent transmission-line monitoring, supporting earlier defect identification and more efficient maintenance planning. The substantial improvement in recall is particularly valuable for safety-critical inspection applications, where missed detections can lead to equipment failures, service interruptions, and increased operational costs.

Future research will focus on real-time deployment aboard UAV platforms, spatiotemporal modeling of inspection videos, and self-supervised pretraining strategies for anomaly-aware feature learning using large-scale unlabeled inspection imagery. Additional investigations will also explore domain adaptation techniques and multimodal inspection systems that combine visual, thermal, and infrared sensing to further improve defect detection robustness under diverse operating conditions.


\begin{thebibliography}{99}

\bibitem{miao2019insulator}
X.~Miao, X.~Liu, J.~Chen, S.~Zhuang, J.~Fan, and H.~Jiang,
``Insulator detection in aerial images for transmission line inspection using
single shot multibox detector,''
\textit{IEEE Access}, vol.~7, pp.~9945--9956, 2019.

\bibitem{li2020insulator}
Y.~Li, X.~Luo, and X.~Yang,
``Insulator defects detection based on deep learning,''
in \textit{Proc. Chin. Autom. Congr. (CAC)}, 2020, pp.~6344--6349.

\bibitem{wang2022self}
Z.~Wang, P.~Liao, C.~Zhong, W.~Deng, L.~Yao, and Z.~Liu,
``Self-explosion defect detection of disc insulators using UAV-based
image processing,''
\textit{IEEE Trans. Instrum. Meas.}, vol.~71, pp.~1--12, 2022.

\bibitem{zhao2021influence}
Y.~Zhao, H.~Zhou, B.~Nie, and X.~Cai,
``Influence of rainfall on the pollution flashover performance of insulators,''
\textit{IEEE Trans. Dielectr. Electr. Insul.}, vol.~28, no.~3,
pp.~921--929, 2021.

\bibitem{ottakath2026flashdetr}
N.~Ottakath, A.~Lutfi, A.~Hamdi, K.~Shaban, and A.~El-Hag,
``FlashDetR: A deep learning pipeline for early detection and time estimation
of flashover in high-voltage insulators using infrared videos,''
\textit{Eng. Appl. Artif. Intell.}, vol.~164, p.~113256, 2026.

\bibitem{saranya2016svm}
K.~Saranya and C.~Muniraj,
``A SVM based condition monitoring of transmission line insulators using PMU
for smart grid environment,''
\textit{Journal of Power and Energy Engineering}, vol.~4, pp.~47--60, 2016.

\bibitem{wang2019image}
J.~Wang, B.~Wen, L.~Li, H.~Zhu, and C.~Luo,
``Insulator detection method in inspection image based on improved faster
R-CNN,''
\textit{Energies}, vol.~12, no.~7, p.~1204, 2019.

\bibitem{elmancy2025real}
A.~Elmancy, A.~Hamdi, K.~Shaban, and A.~El-Hag,
``Real-time insulator defect detection and damage assessment on edge for
UAV-based inspection,''
in \textit{Proc. 26th Int. Middle East Power Syst. Conf. (MEPCON)},
2025, pp.~1--7.

\bibitem{li2021pin}
R.~Li, Y.~Zhang, and D.~Zhai,
``Pin defect detection of transmission line based on improved SSD,''
\textit{High Voltage Engineering}, vol.~47, pp.~3795--3802, 2021.

\bibitem{girshick2014rich}
R.~Girshick, J.~Donahue, T.~Darrell, and J.~Malik,
``Rich feature hierarchies for accurate object detection and semantic
segmentation,''
in \textit{Proc. IEEE Conf. Comput. Vis. Pattern Recognit. (CVPR)},
2014, pp.~580--587.

\bibitem{girshick2015fast}
R.~Girshick,
``Fast R-CNN,''
in \textit{Proc. IEEE Int. Conf. Comput. Vis. (ICCV)},
2015, pp.~1440--1448.

\bibitem{ren2015faster}
S.~Ren, K.~He, R.~Girshick, and J.~Sun,
``Faster R-CNN: Towards real-time object detection with region proposal
networks,''
\textit{IEEE Trans. Pattern Anal. Mach. Intell.}, vol.~39, no.~6,
pp.~1137--1149, 2017.

\bibitem{lu2021insulator}
S.~Lu, J.~Lu, C.~Zhao, W.~Sun, X.~Han, and X.~Zhu,
``Insulator defect detection based on improved Faster-RCNN,''
in \textit{Proc. IEEE Int. Conf. Power Syst. Technol. (POWERCON)},
2021, pp.~1--5.

\bibitem{liao2019power}
G.~P.~Liao, G.~J.~Yang, W.~T.~Tong, W.~Gao, F.~L.~Lv, and D.~Gao,
``Study on power line insulator defect detection via improved Faster
region-based convolutional neural network,''
in \textit{Proc. IEEE 7th Int. Conf. Comput. Sci. Netw. Technol. (ICCSNT)},
Dalian, China, 2019, pp.~262--266.

\bibitem{redmon2016you}
J.~Redmon, S.~Divvala, R.~Girshick, and A.~Farhadi,
``You only look once: Unified, real-time object detection,''
in \textit{Proc. IEEE Conf. Comput. Vis. Pattern Recognit. (CVPR)},
2016, pp.~779--788.

\bibitem{liu2016ssd}
W.~Liu, D.~Anguelov, D.~Erhan, C.~Szegedy, S.~Reed, C.-Y.~Fu, and
A.~C.~Berg,
``SSD: Single shot multibox detector,''
in \textit{Proc. Eur. Conf. Comput. Vis. (ECCV)}, 2016, pp.~21--37.

\bibitem{feng2021electrical}
Z.~Feng, L.~Guo, D.~Huang, and R.~Li,
``Electrical insulator defects detection method based on YOLOv5,''
in \textit{Proc. IEEE 10th Data Driven Control Learn. Syst. Conf. (DDCLS)},
Suzhou, China, 2021, pp.~979--984.

\bibitem{liu2021mti}
C.~Liu, Y.~Wu, J.~Liu, and J.~Han,
``MTI-YOLO: A light-weight and real-time deep neural network for insulator
detection in complex aerial images,''
\textit{Energies}, vol.~14, no.~5, Art.~no.~1426, 2021.

\bibitem{wu2021insulator}
C.~Wu, X.~Ma, X.~Kong, and H.~Zhu,
``Research on insulator defect detection algorithm of transmission line
based on CenterNet,''
\textit{PLoS ONE}, vol.~16, no.~7, Art.~no.~e0255135, 2021.

\bibitem{allam2025attention}
M.~Allam, A.~Hamdi, and S.~Tarek,
``Attention-driven fusion for automated solar panel fault detection in
thermal imagery,''
in \textit{Proc. Int. Conf. Reliable Inf. Commun. Technol. (IRICT)},
2025, pp.~210--220.

\bibitem{elfeky2025automated}
S.~M.~M.~Elfeky, M.~N.~A.~Yehia, and A.~Hamdi,
``Automated MLOps-driven YOLO framework for drone-based plant disease
detection,''
in \textit{Proc. Int. Congr. Inf. Commun. Technol. (ICICT)},
2025, pp.~421--431.

\bibitem{yehia2024enhanced}
M.~N.~A.~Yehia, S.~M.~M.~Elfeky, and A.~Hamdi,
``Enhanced drone-based plant disease detection with visual data augmentation
and optimized training,''
in \textit{Proc. Int. Mobile, Intell., Ubiquitous Comput. Conf. (MIUCC)},
2024, pp.~478--483.

\bibitem{wang2022yolov7}
C.-Y.~Wang, A.~Bochkovskiy, and H.-Y.~M.~Liao,
``YOLOv7: Trainable bag-of-freebies sets new state-of-the-art for real-time
object detectors,''
in \textit{Proc. IEEE/CVF Conf. Comput. Vis. Pattern Recognit. (CVPR)},
2023, pp.~7464--7475.

\bibitem{woo2018cbam}
S.~Woo, J.~Park, J.-Y.~Lee, and I.~S.~Kweon,
``CBAM: Convolutional block attention module,''
in \textit{Proc. Eur. Conf. Comput. Vis. (ECCV)}, 2018, pp.~3--19.

\bibitem{lin2017feature}
T.-Y.~Lin, P.~Doll{\'a}r, R.~Girshick, K.~He, B.~Hariharan, and
S.~Belongie,
``Feature pyramid networks for object detection,''
in \textit{Proc. IEEE Conf. Comput. Vis. Pattern Recognit. (CVPR)},
2017, pp.~2117--2125.

\bibitem{liu2018path}
S.~Liu, L.~Qi, H.~Qin, J.~Shi, and J.~Jia,
``Path aggregation network for instance segmentation,''
in \textit{Proc. IEEE Conf. Comput. Vis. Pattern Recognit. (CVPR)},
2018, pp.~8759--8768.

\bibitem{hu2018squeeze}
J.~Hu, L.~Shen, and G.~Sun,
``Squeeze-and-excitation networks,''
in \textit{Proc. IEEE Conf. Comput. Vis. Pattern Recognit. (CVPR)},
2018, pp.~7132--7141.

\bibitem{elmahdy2024rhrsegnet}
S.~Elmahdy, R.~Hebishy, and A.~Hamdi,
``RHRSegNet: Relighting high-resolution night-time semantic segmentation,''
in \textit{Proc. IEEE Conf. Intell. Methods, Syst., Appl. (IMSA)},
2024.

\bibitem{tan2019efficientnet}
M.~Tan and Q.~V.~Le,
``EfficientNet: Rethinking model scaling for convolutional neural networks,''
in \textit{Proc. Int. Conf. Mach. Learn. (ICML)}, 2019, pp.~6105--6114.

\bibitem{howard2017mobilenets}
A.~G.~Howard, M.~Zhu, B.~Chen, D.~Kalenichenko, W.~Wang, T.~Weyand,
M.~Andreetto, and H.~Adam,
``MobileNets: Efficient convolutional neural networks for mobile vision
applications,''
\textit{arXiv preprint arXiv:1704.04861}, 2017.

\bibitem{sandler2018mobilenetv2}
M.~Sandler, A.~Howard, M.~Zhu, A.~Zhmoginov, and L.-C.~Chen,
``MobileNetV2: Inverted residuals and linear bottlenecks,''
in \textit{Proc. IEEE Conf. Comput. Vis. Pattern Recognit. (CVPR)},
2018, pp.~4510--4520.

\bibitem{tarek2025efficient}
S.~Tarek, A.~Hamdi, and K.~Shaban,
``Efficient segmentation of solar panel defects using knowledge distillation,''
in \textit{Proc. IEEE/ACS 22nd Int. Conf. Comput. Syst. Appl. (AICCSA)},
2025, pp.~1--8.

\bibitem{hinton2006reducing}
G.~E.~Hinton and R.~R.~Salakhutdinov,
``Reducing the dimensionality of data with neural networks,''
\textit{Science}, vol.~313, no.~5786, pp.~504--507, 2006.

\bibitem{lin2017focal}
T.-Y.~Lin, P.~Goyal, R.~Girshick, K.~He, and P.~Doll{\'a}r,
``Focal loss for dense object detection,''
in \textit{Proc. IEEE Int. Conf. Comput. Vis. (ICCV)},
2017, pp.~2980--2988.

\bibitem{zheng2020distance}
Z.~Zheng, P.~Wang, W.~Liu, J.~Li, R.~Ye, and D.~Ren,
``Distance-IoU loss: Faster and better learning for bounding box regression,''
in \textit{Proc. AAAI Conf. Artif. Intell.}, vol.~34, no.~07,
2020, pp.~12993--13000.

\bibitem{zheng2022insulator}
J.~Zheng, H.~Wu, H.~Zhang, Z.~Wang, and W.~Xu,
``Insulator-defect detection algorithm based on improved YOLOv7,''
\textit{Sensors}, vol.~22, no.~22, p.~8801, 2022.

\bibitem{solovyev2021weighted}
R.~Solovyev, W.~Wang, and T.~Gabruseva,
``Weighted boxes fusion: Ensembling boxes from different object detection
models,''
\textit{Image Vis. Comput.}, vol.~107, Art.~no.~104117, 2021.

\end{thebibliography}
\end{document}